%% file: acl2023.tex
\pdfoutput=1

\documentclass[11pt]{article}
\usepackage{authblk}
\usepackage[]{ACL2023}

\usepackage{times}
\usepackage{latexsym}

\usepackage[T1]{fontenc}

\usepackage[utf8]{inputenc}

\usepackage{amsmath}

\usepackage{microtype}

\usepackage{inconsolata}

\usepackage{multirow}
\usepackage{graphicx}
\usepackage{amsmath}
\usepackage[linesnumbered,ruled,vlined]{algorithm2e}
\usepackage{microtype}
\newcommand{\ourbm}{FunGLUE}
\newcommand{\oldbm}{SuperGLUE}
\newcommand{\indophoneme}{Bi-Phone}
\SetKwInput{KwInput}{Input}                
\SetKwInput{KwOutput}{Output}              

\title{\indophoneme: Modeling Inter Language Phonetic Influences in Text}

\author[1]{Abhirut Gupta}
\author[2]{Ananya B. Sai}
\author[1]{Richard Sproat}
\author[1]{Yuri Vasilevski}
\author[1]{\authorcr James S. Ren}
\author[1]{Ambarish Jash}
\author[1]{Sukhdeep S. Sodhi}
\author[1]{Aravindan Raghuveer}
\affil[1]{Google Research}
\affil[2]{IIT Madras}
\affil[ ]{Corresponding author: abhirut@google.com}

\begin{document}
\maketitle
\begin{abstract}
A large number of people are forced to use the Web in a language they have low literacy in due to technology asymmetries. Written text in the second language (L2) from such users often contains a large number of errors that are influenced by their native language (L1).
We propose a method to mine phoneme confusions (sounds in L2 that an L1 speaker is likely to conflate) for pairs of L1 and L2.
These confusions are then plugged into a generative model (\indophoneme) for synthetically producing corrupted L2 text.
Through human evaluations, we show that \indophoneme~generates plausible corruptions that differ across L1s and also have widespread coverage on the Web.
We also corrupt the popular language understanding benchmark \oldbm~with our technique (\ourbm~for Phonetically Noised GLUE) and show that SoTA language understating models perform poorly.
We also introduce a new phoneme prediction  pre-training task which helps byte models to recover performance close to \oldbm. Finally, we also release the \ourbm~benchmark to promote further research in phonetically robust language models. To the best of our knowledge, \ourbm~is the first benchmark to introduce L1-L2 interactions in text.  

\end{abstract}

\input{intro}

\input{related_work}

\input{method}
\input{rtt}
\input{biphone}

\input{human_eval}

\input{coverage_analysis}
\input{super_clue}

\input{robust_models}
\section{Conclusion} 
Language is a significant barrier to technology especially for new internet users. For such users, English often is not their first language. The speech community has made significant progress in making technology (ASR for instance) accessible for such users by making models robust to account for inter-language interactions. We argue that a similar line of effort is needed in the Natural Language Understanding for Text community as well. To this end, we first propose a generative model {\em \indophoneme} that can account for L1-L2 interactions in text. Next we show the inter-language perturbations generated by \indophoneme~are indeed present in non-trival amount in the common crawl corpus. We also release a new benchmark~\ourbm~to help further research in this area. We also present our early yet very promising explorations on making natural language understanding models robust to L1-L2 phonetic shifts through a novel phoneme prediction based pre-training.

\section{Limitations} 
\textbf{Algorithmic Limitations:} The current approach assumes each phoneme / grapheme corruption is independent of the surrounding phonemes / graphemes, which can be relaxed to get further insights and model any contextual phonetic shifts. The relative importance between grapheme and phoneme corruptions could also be explored as a hyperparameter to personalize more to the type of errors of a community.\\
\textbf{Other Limitations} (with respect to available data and existing resources): Our coverage analysis is conservative since it does not cover the user generated data from various social media where such L1-L2 phonetic misspellings are bound to be more common. The coverage analysis also relies on the context not being corrupted. However, this might not necessarily hold and the analysis could benefit from a careful formulation of a relaxed matching criteria that also considers cases with corrupted contexts.
With transliteration playing a major role in our solution, it is difficult to immediately extend the work to low-resource languages that do not have models or appropriate datasets to build transliteration modules.  
\bibliography{anthology,custom,l2refs}
\bibliographystyle{acl_natbib}






\end{document}

%% file: intro.tex
\section{Introduction}
\label{Intro}

We live in a multilingual world with over 7,000 languages spoken across the globe \cite{ethnologue}. However, technology asymmetrically supports only a 
few specific languages. For instance, the internet is mostly in English with over 60\% of websites using the language despite just around 16\% share of its speaking population around the world\footnote{https://w3techs.com/technologies/overview/content\_language} \cite{DBLP:journals/corr/cs-CL-0006032_eng_on_www_web}.  
Increasingly, people are forced to navigate and produce content on the web in languages they have not been formally trained on.  The English text produced by ESL (English as Second / L2 language) writers is heavily influenced by their native language (L1). 

Research in the field of second-language acquisition has found evidence of phoneme-shift based misspellings stemming from L1 influence in L2 text for specific language pairs~\cite{Ibrahim:78,Cook:97,Bestgen:Granger:11,Sari:15,Ogneva:18,Motohashi:Ishizawa:20}.
Studies in Natural Language Understanding (NLU) have been limited to spelling correction~\citet{NAGATA2017474,flor-etal-2019-benchmark} and native language identification~\citet{DBLP:conf/acl/ChenSN17_nli,DBLP:conf/bea/NicolaiHSYK13_cognate} in English learners. These studies predominantly use the TOEFL11 dataset \cite{article_toefl11_data} which deals with very specific demographics such as test-takers who have formal training in the L2 language.


We make the following four key observations about prior work in the study of  L1-L2 influences in text and speech. First, current models for  L1-L2 influence on textual spelling are limited to certain language pairs and tasks. We argue that L1-L2 influence phenomenon is much more broad and is language and task agnostic. Second, there is no large scale study to examine the prevalence of this phenomenon on the open web. Third, given that this is an important problem especially for multi-lingual, new-to-the-internet communities there is no standardized benchmark to study the robustness of natural language understanding(NLU) and Natural Language Generation (NLG) models to inter-language phonetic noise. Finally, there is very sparse literature on architecture / pre-training strategies to introduce phonetic robustness into large language models.  In this paper, we present modeling techniques,data analyses and a new benchmark to address the gaps mentioned above. We summarise our contributions as follows:

\begin{enumerate}
\item We propose a language-agnostic method to mine phoneme confusions that arise due to interference between a native language (L1) and second language (L2).  Our method exploits the ``hidden knowledge" contained in $L1 \rightarrow L2$ and $L2 \rightarrow L1$ transliteration models. 
We also propose a generative model {\em Bi-Phone} that is able to synthetically produce spelling corruption in accordance with L1-L2 confusions (Sections~\ref{sec:rtt},~\ref{sec:biphone}).  
\item  Through human evaluation and coverage analysis we show that {\em Bi-Phone} produces  spelling corruptions that are not only deemed plausible by native L1 speakers but also have substantial coverage in the open web crawl corpus. To the best of our knowledge no prior work has demonstrated the presence of L1-L2 phonetic corruptions in a large scale, common dataset like Common Crawl (Section~\ref{sec:eval}). 
\item We release a dataset consisting of sentences with L1-L2 phonetic spelling corruptions found in Common Crawl. We also release a benchmark called \ourbm, an extension of the~\oldbm~benchmark for L1-L2 spelling corruptions. To the best of our knowledge \ourbm~is the first benchmark to measure the robustness of models to L1-L2 interference in text (Section~\ref{sec:superCLUE}). 
\item We show SoTA models do not perform well on \ourbm. We then introduce a novel pre-training task of phoneme prediction, which together with byte level architectures substantially bridges the gap on the noised benchmark (by up to 11\% absolute on certain test sets). This is particularly impressive since this gain is achieved without ever showing the model any noised examples (Section~\ref{sec:robust_models}). 
\end{enumerate}

%% file: related_work.tex
\section{Related Work}

We divide the presentation of related work in two sections. (i) First, we discuss prior work spanning multiple research areas regarding phonetic influences in text and how it relates to our work. (ii) Second, we discuss work in the speech domain which studies phonetic variations occurring due to inter-language interference in multi-lingual scenarios. 

\subsection{Phonetic Influences in Text}
Phonetic influence on spelling errors has been studied in the past~\cite{10.1145/146370.146380,toutanova-moore-2002-pronunciation,electronics9101670_survey_automatic_spell_corr}. The source of such errors is that both native and non-native speakers resort to phonetic spellings for unfamiliar words or names. This direction of work does not address the effect of native language (L1) based phoneme shifts on second-language (L2) spellings.

There has also been work that focuses on learner English~\footnote{learner English refers to English as a foreign language} for different applications. ~\citet{NAGATA2017474,flor-etal-2019-benchmark} study automatic spell correction with distributional methods that require a larger learner corpus. ~\citet{DBLP:conf/acl/ChenSN17_nli,DBLP:conf/bea/NicolaiHSYK13_cognate} explore Native Language Identification (NLI) on such text. 
A widely used dataset for these learner English tasks is the TOEFL11 corpus \cite{article_toefl11_data} which contains English essays written by non-native test-takers.
It is important to note that these analysis are limited to misspellings made by authors with sufficient L2 knowledge/ training that qualifies them to take the test.
They also do not explicitly study the causes of the misspellings or the inter-language interference.

There has also been a fair amount of interest in the second-language acquisition field on the influence of L1 on L2 spelling. ~\citet{Ibrahim:78,Cook:97,Bestgen:Granger:11,Sari:15,Ogneva:18,Motohashi:Ishizawa:20} all find evidence of such influence in specific language pairs. These often stem from the lack of certain sounds in L1 leading to difficulty in distinguishing similar sounds in L2. They also find more interesting phenomenon like L1 constraints on consonant clusters are reflected in L2 spellings by learners. While this direction of research is highly pertinent to our work, our goal is to generate plausible L1-L2 phonetic shift based misspellings more generally instead of studying the phenomenon in particular language pairs.

\subsection{Inter-language Influence for Phonetic Deviations in Speech} 
Phonetic variations of words have been well-studied in the context of speech applications. Several studies \cite{DBLP:journals/ejasmp/RadzikowskiNWY19_dual_supervised,DBLP:journals/corr/abs-2011-06226_spoken_term_detect,DBLP:journals/ejasmp/RadzikowskiWYN21, DBLP:conf/petra/BirdWEF19_accent_class} discuss the drop in performance of speech applications such as ASR, spoken-term detection, etc., when presented with non-native speech data. 
They attribute this drop mainly to the nuances in pronunciation that are often not present in the training data, due to the lack of sufficient non-native speech data. 
To address and close this gap, several strategies ranging from the use of cross-lingual/multi-lingual phonological inventories to end-to-end training have been applied.
However, these studies do not focus on how the same phonetic influences manifest in written text.

%% file: method.tex
\section{Method}
\label{sec:method}
In this section we introduce our method for creating inter-language influenced phonetic misspellings (or corruptions). We present the technique in two parts. Section~\ref{sec:rtt} presents a method for mining native-language influenced phonetic confusions. Section~\ref{sec:biphone} contains details of Bi-Phone, our model that uses mined phonetic confusions to create misspellings.

%% file: rtt.tex
\subsection{Mining Phoneme-Phoneme Confusions}\label{sec:rtt}
\begin{figure}[t]
    \centering
    \includegraphics[width=\linewidth]{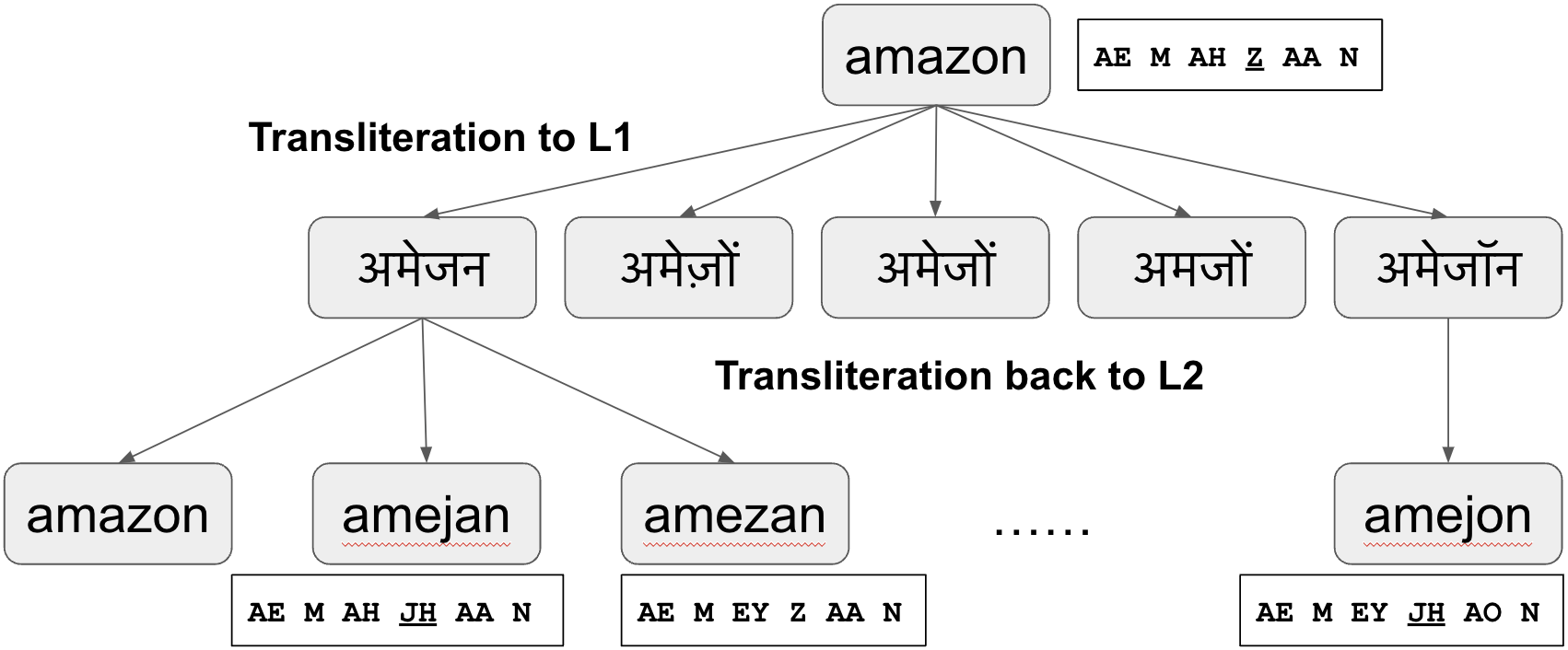}
    \caption{Overview of the Round Trip Transliteration method for creating word pairs from which phoneme confusions are mined. In this example, we create pairs for the dictionary word ``amazon" with round-trip transliteration through Hindi as the pivot language. Phoneme sequences for the original and round-trip transliterated words are also shown. Multiple words with \textit{JH} in the round-trip transliterations enables us to map the \textit{Z} sound to the \textit{JH} sound for Hindi speakers.}
    \label{fig:rtt}
\end{figure}
\begin{figure}[t]
    \centering
    \includegraphics[width=\linewidth]{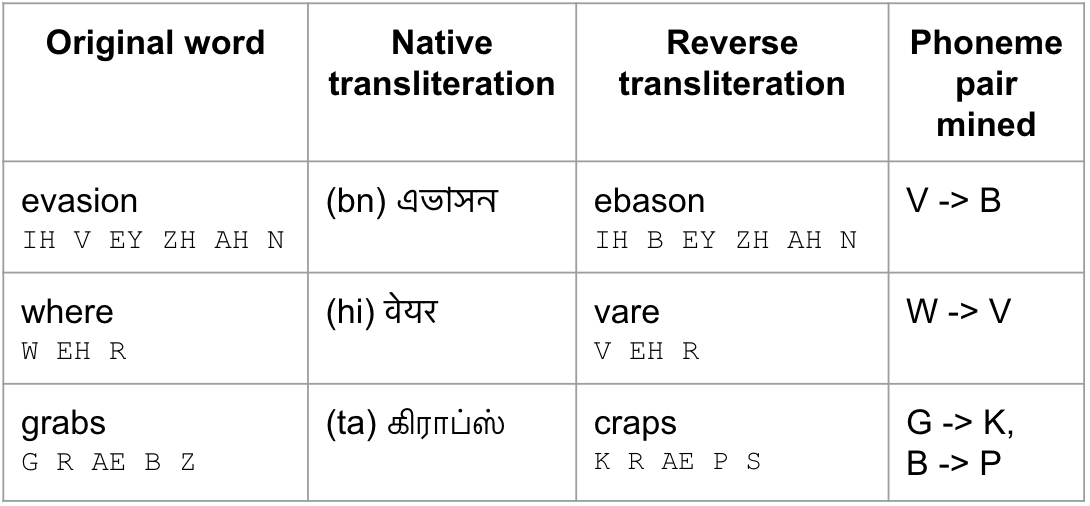}
    \caption{Examples of round trip transliterations of dictionary words with different pivot languages, the corresponding phoneme sequences, and the phoneme confusion mined. While the third example also has a Z -> S shift, it is not mined because we only consider the top-10 most frequent confusions per (L1, L2) pair.}
    \label{fig:rtt_example}
\end{figure}
The first problem is to identify possible phoneme confusions that a speaker of a given native language (L1) is likely to encounter when speaking a second language (L2). These confusions can be imagined as a matrix $C(L1, L2)$, which contains likelihood of the $i_{th}$ L2 phoneme ($ph_i$) being confused as the $j_{th}$ L2 phoneme ($ph_j$) by a native speaker of L1 as the value in the cell $C(L1, L2)[i][j]$.
\begin{equation}
    C(L1, L2)[i][j] = P(ph_j | ph_i)
\end{equation}
Building this matrix across all pairs of languages is an expensive task.
It is also challenging to accurately determine the likelihood of such confusions without large datasets of parallel words.

Transliteration models are trained on large parallel datasets with the objective of transcribing sounds representing words in one language with in the script of a different language. 
They imbibe important information about sounds in one language that are indistinguishable in another (and therefore lexicalized identically). 
We propose a round-trip transliteration based method which aims to mine these phoneme confusions and their likelihoods from this knowledge hidden in transliteration models. We collect a large dictionary of English words (our chosen L2) and apply two steps of transliteration~\footnote{https://github.com/libindic/indic-trans}~\cite{Bhat:2014:ISS:2824864.2824872_rtt} to convert them back to English via a pivot language (L1), as shown in Figure~\ref{fig:rtt}. We then align the phoneme sequence of the original word with that of its round-trip transliterated version using the Needleman-Wunsch algorithm \cite{NEEDLEMAN1970443}. We count the frequency of each of the possible sound-shifts in the whole corpus to estimate likelihood.
Figure~\ref{fig:rtt_example} shows examples of word pairs created through different pivot languages and the phoneme confusion mined from these. We consider only the top-10 most frequent phoneme confusions per (L1, L2) for the next step.


%% file: biphone.tex
\subsection{BiPhone: A Generative Model for L1-L2 Phonetic  Misspellings}\label{sec:biphone} 
The second problem we focus on is to create a model for sampling phonetic misspellings ($\boldsymbol{\tilde{w}}$) for a given word ($\boldsymbol{w}$) in L2 that a native speaker of L1 is likely to make. We can represent the probability distribution learnt by this model as $P(\boldsymbol{\tilde{w}} | \boldsymbol{w})$. Assuming a deterministic mapping from the word $\boldsymbol{w}$ to its phoneme sequence $\boldsymbol{ph_w}$, and introducing the corrupted phoneme sequence ($\boldsymbol{ph_{\tilde{w}}}$) that finally generates $\boldsymbol{\tilde{w}}$, we can rewrite it as -
\begin{equation}
\begin{split} 
    P(\boldsymbol{\tilde{w}}| \boldsymbol{w}) &
    = P(\boldsymbol{\tilde{w}} | \boldsymbol{ph_w})   \\
   &  = \sum_{\boldsymbol{ph_{\tilde{w}}}} P(\boldsymbol{ph_{\tilde{w}}} | \boldsymbol{ph_w}) * P(\boldsymbol{\tilde{w}} | \boldsymbol{ph_{\tilde{w}}})
\end{split} 
\label{eq:gen_model}
\end{equation}
Here a word $\boldsymbol{w}$ is comprised of graphemes $\{w^1, w^2, ..\}$ where $w^i \in Graphemes(L2)$ and a phoneme sequence $\boldsymbol{ph_w}$ is comprised of phonemes $\{ph^1, ph^2, ..\}$ where each individual phoneme $ph^i$ is from the set of available phonemes for $L2$. In our experiments, we use the ARPAbet phoneme set for English~\footnote{https://en.wikipedia.org/wiki/ARPABET}.

\textbf{Phoneme-Phoneme Error Model}: The first term under the summation in Equation~\ref{eq:gen_model} models the likelihood of generating a corrupted phoneme sequence $\boldsymbol{ph_{\tilde{w}}}$ given that a native speaker of 
L1 is attempting to speak a phoneme sequence $\boldsymbol{ph_w}$ in L2. With simplifying independence assumptions that each phoneme is corrupted individually, independent of phonemes around it, we can factorize this term to utilize the phoneme confusion matrix we have mined.
\begin{equation}
\begin{split}
    \hspace{2em}&\hspace{-2em}P(\boldsymbol{ph_{\tilde{w}}} | \boldsymbol{ph_w}) =\prod_{i} P(ph_{\tilde{w}}^i | ph_{w}^i) \\
     &=\prod_{i} C(L1, L2)[ph_{w}^i][ph_{\tilde{w}}^i]
\end{split} 
\label{eq:ph_ph_error_model}
\end{equation}

\textbf{Phoneme-Grapheme Density Model}: The second term in Equation~\ref{eq:gen_model} expresses the probability of generating the grapheme sequence to represent $\boldsymbol{\tilde{w}}$ given the phoneme sequence $\boldsymbol{ph_{\tilde{w}}}$. 
We can assume equal lengths for the two sequences, by allowing some phonemes to not generate any graphemes, when necessary.
Again, we make independence assumptions where the grapheme used to represent a given phoneme does not depend on neighbouring phonemes or graphemes.
\begin{equation}
P(\boldsymbol{\tilde{w}} | \boldsymbol{ph_{\tilde{w}}}) =\prod_{i} P(\tilde{w}^i | ph_{\tilde{w}}^i)
\label{eq:ph_gh_error_model}
\end{equation}
To compute $P(\tilde{w}^i | ph_{\tilde{w}}^i)$, we use a pronunciation dictionary in L2 (CMUDict\footnote{http://www.speech.cs.cmu.edu/cgi-bin/cmudict} for English).
First, phoneme-character probabilities are generated through alignment. Next, for each word, character sequences are converted to graphemes by maximizing the alignment score. Finally, the various phoneme-grapheme alignments along with their frequencies are converted to probabilities by dividing it by the frequency of the phoneme.

\textbf{Inference}: Given an original phoneme sequence for a word to be corrupted, we begin sampling with a fixed width (K) beam from left to right. At each position, we pick the top-K candidates comprising both phoneme-phoneme shifts and phoneme-grapheme alternatives greedily. Since both Phoneme-Phoneme Error Model and Phoneme-Grapheme Density Model are context independent, the greedy strategy gives us the global top-K misspellings. Identity corruptions are removed as a final step.



%% file: human_eval.tex
\section{Evaluations}\label{sec:eval} 
\begin{table}
\small
\begin{center}
\begin{tabular}{|c|c|c|c|}
\hline
\textbf{Phoneme Shift} & \textbf{Hi}  & \textbf{Ta} & \textbf{Bn} \\ \hline
AH2 -> AH0 & 100\% & - & 100\% \\ \hline
IH2 -> IH0 & 100\% & - & 100\% \\ \hline
ER2 -> ER0 & 100\% & - & -\\ \hline
DH -> TH & 54\% & - & 62\% \\ \hline
ER2 -> ER0 & 95\% & - & -\\ \hline
D -> T & - & 30\% & - \\ \hline
B -> P & - &39\% & - \\ \hline
DH -> D & - & 0\% & - \\ \hline
G -> K & - & 47\% & -\\ \hline
V -> B & - & - & 58\% \\ \hline
Z -> S & - & - & 50\% \\ \hline
\end{tabular}
\end{center}
\caption{Plausibility scores for different phoneme shifts across Hindi, Tamil, and Bengali.}
\label{tab:human_eval}
\end{table}
\begin{table}
\resizebox{1\linewidth}{!}{
\begin{tabular}{|c|c|c|c|}
\hline
L1 & Correct   & Misspelt   & Phoneme  \\
  &Word & Word &  Variation \\
\hline
Hindi & they &	thay & DH -> TH \\\hline
Tamil & exam	 & eksam  & G -> K\\ 
&  bacterial &	pactirial & B -> P\\ \hline 
Bengali & very &	bery  & V -> B \\ 
 & equation &	ikvasan & ZH -> S \\ \hline
\end{tabular}
}
\caption{Examples of highly plausible misspellings as rated by native speakers for various L1 languages with L2 language as English}
\label{tab:human_eval_examples}
\end{table}
We evaluate the misspellings generated by our model along two distinct dimensions.
\subsection{Plausibility}
For evaluating plausibility of generated misspellings from~\indophoneme, we focus on three native languages (L1) : Hindi, Tamil and Bengali with English as the non-native language (L2). Hindi and Bengali are the two most widely spoken languages in India and among the top few in the world. Tamil is also a widely spoken language in India and introduces typological diversity in our analysis. Finally, our choice of L1 is also based on availability of native speakers for the annotation task.

For each language, we present 150 randomly selected word, misspelling pairs generated from~\indophoneme~to native speakers (5 for Hindi, 3 for Tamil and Bengali each). Rater instructions are as follows:  
Given a list of pairs in English (correct word, misspelling), the task is to evaluate if the misspelling is plausible for pronunciation shifts often made by speakers of the given first language.
For example - Bengali speakers often shift the “v” sound to “b” so, “evicted” could be plausibly misspelt as “ebicted” or “abicted”. Each rater provides a 1 or 0 to indicate whether the variant looks plausible or not, respectively. We use a simple majority to assign an overall label to each pair. The raters for this task are our colleagues who are native speakers of the language they are annotating for.


Table~\ref{tab:human_eval} reports the percentage of misspellings rated as plausible for each phoneme shift. We observe that misspellings for Tamil are rated as less plausible than for other languages. The reason for this is the more drastic phoneme shifts uncovered in Tamil (B -> P and G -> K). However, misspellings stemming from these shifts are still not rated as completely implausible, which emphasizes that these shifts are indeed common. We also measure inter-annotator agreement through kappa scores which are 0.40 for Hindi, 0.37 for Tamil, and 0.34 for Bengali.


%% file: coverage_analysis.tex
\subsection{Prevalence: Coverage Analysis}
\label{sec:coverage_of_biphone}
\begin{figure}[t]
    \centering
    \includegraphics[width=0.8\linewidth]{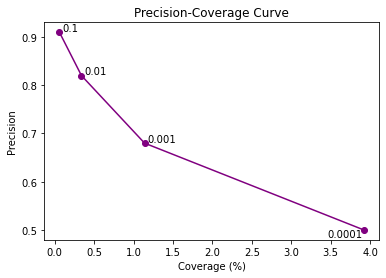}
    \caption{Precision and coverage plotted at different misspelling confidence scores (labels on points). Coverage is represented as a fraction of 31,755,066 sentences that have atleast one non-English dictionary word.}
    \label{plt:coverage_pr_cov}
\end{figure}
In the previous section we investigate the plausibility of phoneme-shifts mined by~\indophoneme~and the misspellings created as a result. However, this investigation does not throw light on the pervasiveness of such misspellings in real world content.

In this section, we aim to evaluate the severity of the phonetic misspelling issue by uncovering such misspellings in web data. For our analysis, we use the Common Crawl\footnote{\url{https://commoncrawl.org/}} corpus, which is a publicly available scrape of real web data. While most existing language work deals with a highly cleaned version of this corpus~\cite{raffel2020exploring}, we skip such filtering and cleaning steps to retain noisy, user-generated text. We only use Hindi as the native language (L1) in this analysis. Our analysis has three distinct steps - (1) Candidate Sentence Retrieval, (2) Misspelling Confidence Scoring, and (3) Human Evaluation.

\textbf{1. Candidate Sentence Retrieval:} We begin our analysis by creating 10 misspellings of the top 10,000 most common English words from the Google ngram corpus~\cite{michel2011quantitative} and words that make up 90\%-ile of the English words in the Common Crawl corpus. Our hypothesis is that the most common words in English are also the most likely to be misspelt with native language influences. Our pool of sentences is the set of all sentences with at least one non-English dictionary word. The size of this pool is 31,755,066 sentences. From this pool, we create our candidate set by retrieving all sentences that contain one of our generated misspellings.

\textbf{2. Misspelling Confidence Scoring:} The next step is to ascertain that the misspellings retrieved are indeed a noisy form of the intended original word and not a completely different word. For example, ``vare" could be a corruption of the English word ``where" with the W -> V sound shift, or it could be the less used English word meaning a weasel~\footnote{https://www.merriam-webster.com/dictionary/vare}. We use a simple 1-word left and right context for this disambiguation. For every occurrence of a potentially misspelt word $\hat{W}$ in context $(L_{\hat{W}}, \hat{W}, R_{\hat{W}})$, we evaluate the probability of seeing the corresponding clean word ($W$) in the same context. This likelihood, $P(L_{\hat{W}}, W, R_{\hat{W}})$ computed as follows can be used as a score to represent our confidence in the retrieved misspelling.
\resizebox{\columnwidth}{!}{
\begin{minipage}{\linewidth}
\begin{align*}
    &P(L_{\hat{W}}, W, R_{\hat{W}}) \\
    &= \frac{F(L_{\hat{W}}, W, R_{\hat{W}})}{\sum_{w}F(L_{\hat{W}}, w, R_{\hat{W}})} 
   \text{\ \  , \ \ if} \sum_{w}F(L_{\hat{W}}, w, R_{\hat{W}}) > 0 \\
   &= 0.4 * \Bigg[\frac{F(L_{\hat{W}}, W)}{\sum_{w}F(L_{\hat{W}}, w)}
   + \frac{F(W, R_{\hat{W}})}{\sum_{w}F(w, R_{\hat{W}})} \Bigg] 
    \text{, otherwise}
\end{align*}
\label{eq:scoring}
\end{minipage}
}

Here 0.4 is the backoff-weight following the \textit{Stupid Backoff} technique from~\citet{brants-etal-2007-large}.

We can compute the coverage of~\indophoneme~ in web data by considering the fraction of sentences where the misspelling confidence score is greater than a certain threshold over the total number of sentences in our original pool.

\textbf{3. Human Evaluation:} Finally, we also sample a subset of the sentences to have human raters verify that our retrieved misspellings indeed correspond to the original word. We show raters the original retrieved sentence which contains the generated misspelling and a parallel sentence where the misspelling has been replaced with the original word and ask raters if this correction is valid in the given context. We can compute a reliable metric for precision with this human evaluation. Ratings for this task are fetched from a cloud rating service where raters are bilingual Hindi-English speakers with a graduate degree.

Figure~\ref{plt:coverage_pr_cov} presents the precision and coverage at different thresholds of misspelling confidence score. At threshold 0.001, we have roughly 70\% precision while still having a coverage of ~1.14\% (362,472 sentences*). The size of the initial pool (30 million candidate sentences) and the simple method used for our analysis underline how prevalent such misspellings are. Also it is important note that such misspellings will be even more prevalent in a purely UGC (user generated content) corpus. C4 contains a significant fraction of clean English web pages.

%% file: super_clue.tex
\section{The~\ourbm~Benchmark}\label{sec:superCLUE}
\begin{table}
\small
\begin{center}
\begin{tabular}{|p{0.07\linewidth}|p{0.6\linewidth}|p{0.17\linewidth}|}
\hline
\textbf{Split} & \textbf{Description}  & \textbf{Contains Phonetic Noise} \\ \hline
train & Train split from SuperGLUE as is & No\\\hline
dev & Dev split from SuperGLUE as is & No\\\hline
test & Dev split from SuperGLUE noised with BiPhone & Yes\\\hline
\end{tabular}
\end{center}
\caption{Description of splits in \ourbm. Checkpoint selection is done on the dev set which does not contain phonetic misspellings. The test set is used only for reporting results.}
\label{tab:super_clue_splits}
\end{table}

\begin{table}
\small
\begin{center}
\begin{tabular}{|c|c|}
\hline
Task & Field Name \\ \hline
BoolQ & question \\
CB & premise \\
COPA & premise \\
MultiRC & question \\
ReCoRD & query \\
RTE & hypothesis \\
WiC & sentence1 \\\hline
\end{tabular}
\end{center}
\caption{Fields we noise for different task when creating~\ourbm.}
\label{tab:super_clue_noised_fields}
\end{table}

Significant progress has been made in recent research to substantially improve performance of language understanding tasks. 
~\oldbm~\cite{wang2019superglue} is a very popular benchmark with ten diverse and hard language understanding tasks. These tasks are BoolQ, CommitmentBank (CB), Multi-Sentence Reading Comprehension (MultiRC), Choice of Plausible Alternatives (COPA), Reading Comprehension with Commonsense Reasoning (ReCoRD), Recognizing Textual Entailment (RTE), Words in Context (WiC), Broadcoverage Diagnostics (AX-b), The Winograd Schema Challenge (WSC), and Winogender Schema Diagnostics (AX-g). We argue that for language understanding models to be effective for bi-lingual users, they must be robust to inter-language phonetic spelling variations.
Towards this end, we introduce~\ourbm\footnote{https://github.com/google-research-datasets/FunGLUE} which stands for Ph(F)onetically noised GLUE where randomly selected words from tasks in the~\oldbm~benchmark are corrupted with Bi-Phone based misspellings. It is extremely important to note that we only create a hold-out evaluation set created by introducing misspellings to the SuperGLUE development set. The training set is left clean to mimic real world scenarios where noised training data is difficult to obtain. Additionally, it would be unfair to train and evaluate models on synthetic misspellings from the same source. Table~\ref{tab:super_clue_splits} summarizes the training, validation, and test sets in \ourbm.

Misspellings for words in the original task are created from~\indophoneme~with the following design choices:

\textbf{(i) What to noise:} Since we want to keep the task realistic, we only introduce misspellings in certain pre-selected fields and not all text fields. This reflects real world situations where content is often available in well spelt English but user queries have phonetic errors. Table~\ref{tab:super_clue_noised_fields} presents the fields we actually noise.

\textbf{(ii) Which misspellings to use:} Since we expect benchmarks to have a high quality, we put in a number of guardrails to ensure poor quality misspellings do not make it through to the benchmark. First, we only use~\indophoneme~misspellings with Hindi and Bengali as native language since Tamil misspellings were rated as less plausible by native speakers. Next, we noticed that plausibility scores drop for words smaller than 4 characters, so we only noise longer words. We also filter out misspellings that contain certain patterns of implausible noise generated by our Grapheme2Phoneme model with rules. Finally, all (word, misspelling) pairs used in~\ourbm~are manually verified by members of the team as plausible.

\begin{table}
\small
\begin{center}
\begin{tabular}{|c|c|c|}
\hline
\textbf{Task} & \textbf{Tokens misspelt}  & \textbf{Examples w/ noise} \\ \hline
boolq & 30.6\% & 96.2\%\\\hline
cb & 29.5\% & 96.4\% \\\hline
multirc & 33.8\% & 96.4\% \\\hline
copa & 25.2\% & 78.0\% \\\hline
record & 29.5\% & 99.4\% \\\hline
rte & 35.9\% & 97.1\% \\\hline
wic & 28.9\% & 84.0\% \\\hline
\end{tabular}
\end{center}
\caption{Stats on amount of noise added in~\ourbm.}
\label{tab:super_clue_stats}
\end{table}

\textbf{(iii) How much noise to add:} Since we do not want to artificially introduce too much noise, we only replace ~30\% of words from the original benchmark across tasks. Table~\ref{tab:super_clue_stats} contains stats on the amount of noise added to each task. We were currently unable to include the noised version of the WSC, AX-b and AX-g tasks due to some difficulties in accessing the eval sets. We plan to include this with the final data release.

\subsection{Models}
\begin{figure}[t]
    \centering
    \includegraphics[width=0.9\linewidth]{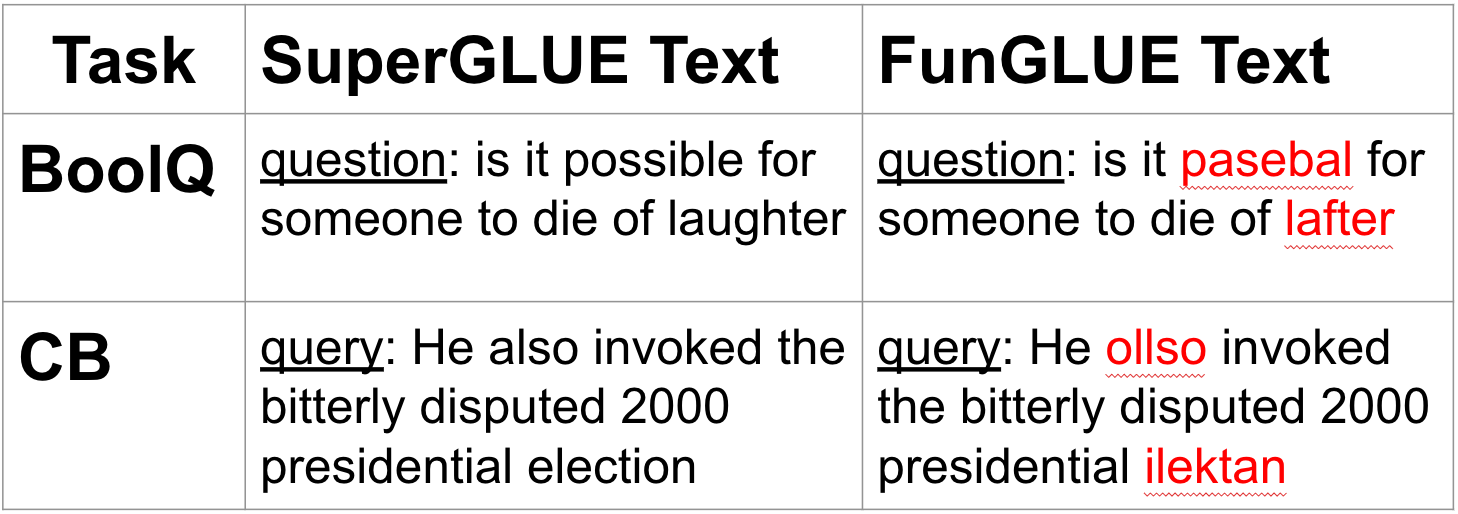}
    \caption{Examples from validation set of two tasks in~\ourbm~against~\oldbm. Words which are replaced with their noised versions are in red.}
    \label{fig:super_clue_example}
\end{figure}
In this section we investigate if state-of-the-art models are robust to the phonetic noise introduced by~\ourbm  ~by comparing their performance on~\oldbm. 
For this purpose, we consider mT5 \cite{DBLP:conf/naacl/XueCRKASBR21_mt5} and ByT5 \cite{DBLP:journals/corr/abs-2105-13626_byt5} models. These are both transformer based sequence-to-sequence models that frame all language understanding tasks as sequence generation.
mT5 uses sub-word tokenization built on a multilingual corpus, to represent text. It should therefore be more robust to input variations than comparable models with tokenization on monolingual corpora with lower diversity. ByT5 avoids the tokenization step by building input representations from individual bytes, and is designed to perform more gracefully on noisy text across a range of tasks.

For all models, we use the base architecture. Since training these models is expensive, we do not perform any hyper-parameter search. Instead, we use fine-tuning parameter values from the original papers. Crucially, fine-tuning for all models is performed identically on clean data from~\oldbm. We use the same mixture of tasks as in~\citet{2020t5}. Fine-tuning is done for up to 200,000 steps and the best checkpoint is picked based on performance on the clean dev set from~\oldbm. We use 16 TPUv3s for fine-tuning all models.

\subsection{Spell Correction Baselines}
Spell correction methods provide obvious baselines when dealing with incorrectly spelt data. Spell corrected data can then be use to run inference with existing models. To evaluate the merit of this technique, we measure performance after correction from two state of the art approaches: (1) NeuSpell BERT~\cite{jayanthi-etal-2020-neuspell} - spell corrector built on top of BERT. (2) BERT-Large mask prediction - using a BERT Large model for predicting the correct word in positions where we have misspellings. In both of these approaches, we provide the positions of incorrectly spelt words. This is an advantage since this information is not available in real world noisy text. We compare the performance of both mT5 and ByT5 on \ourbm~eval sets corrected by these approaches.

\subsection{Results}
\begin{table*}
\small
\begin{center}
\resizebox{1\textwidth}{!}{
\begin{tabular}{|p{0.02\textwidth}|l|c|c|c|c|c|c|c|c|c|c|}
\hline
\multirow{2}{*}{No.}& \multirow{2}{*}{Model} & BoolQ & \multicolumn{2}{c|}{CB} & COPA & \multicolumn{2}{c|}{MultiRC} & \multicolumn{2}{c|}{ReCoRD} & RTE & WiC \\ \cline{3-12}
& & Acc & Acc & F1 & Acc & EM & F1 & EM & F1 & Acc & Acc \\ \hline
\multicolumn{12}{c}{\oldbm} \\ \hline
1 & mT5 & 78.10 & 92.86 & 90.53 & 61.00 & 33.68 & 73.03 & 67.22 & 68.26 & 74.37 & 68.03\\ \hline
2 & ByT5 & 79.20 & 91.07 & 90.37 & 58.00 & 32.00 & 70.14 & 72.10 & 72.79 & 81.23 & 70.85 \\ \hline

\multicolumn{12}{c}{\ourbm} \\ \hline
3 & mT5 & 68.81 & 80.36 & 74.21 & 55.00 & 28.23 & 70.37 & 58.46 & 59.46 & 67.87 & 63.64\\ \hline

3a & mT5 - NeuSpell & 67.92 & 76.79 & 74.99 & 64.00 & 30.43 & 70.85 & 60.36 & 61.33 & 65.34 & 65.83 \\ \hline
3b & mT5 - Bert-L mask pred & 66.42 & 71.43 & 79.6 & 57.00 & 27.70 & 67.91 & 55.6 & 56.63 & 58.84 & 62.54 \\ \hline

4 & ByT5 & 74.04 & 80.36 & 73.67 & 58.00 & 32.42 & 72.73 & 67.54 & 68.19 & 70.40 & 66.46 \\ \hline

4a & ByT5 - NeuSpell & 72.84 & 76.79 & 67.86 & 54.00 & 32.53 & 72.47 & 63.64 & 64.25 & 69.68 & 66.46 \\ \hline
4b & ByT5 - Bert-L mask pred & 70.52 & 75.00 & 70.7 & 55.00 & 26.76 & 68.60 & 59.75 & 60.35 & 64.62 & 64.26 \\ \hline


5 & Phonetic mT5 & 71.80 & 80.36 & 73.66 & 53.00 & 25.81 & 72.2 & 55.85 & 56.86 & 61.37 & 63.17\\ \hline
6 & Phonetic ByT5 & 74.37 & 87.50 & 85.46 & 66.00 & 33.26 & 75.15 & 70.21 & 70.88 & 76.17 & 66.77 \\ \hline
\end{tabular}
}
\end{center}
\caption{\textbf{First 4 rows}: Performance of SoTA models on tasks in the~\oldbm~and~\ourbm~(noised) benchmarks. Performance of both mT5 and ByT5 (rows 3 and 4 compared to 1 and 2) drops on the noised benchmark, although ByT5 (row 4) is slightly more robust.
\textbf{Rows 3a, 3b, 4a, and 4b} show the performance of mT5 and ByT5 after misspelt words in the eval set are replaced with corrections from SoTA techniques. While mT5 benefits slightly from such corrections, ByT5 performance is worse across all tasks after spell correction is applied. This demonstrates the inability of current spell correction models to handle such misspellings.
Rows 3a and 4a correspond to corrections from the NeuSpell~\cite{jayanthi-etal-2020-neuspell} model. Rows 3b and 4b correspond to corrections using mask prediction from a Bert-Large model.
\textbf{Last 2 rows}: Performance of the same models when trained on a few additional steps with the phoneme prediction task on clean data (Phonetic mT5 and ByT5). The ByT5 (row 6 compared to row 4) model gains substantially with such pre-training.}
\label{tab:super_gloo_results}
\end{table*}
Rows 1-4 in Table~\ref{tab:super_gloo_results} show the performance of mT5 and ByT5 on~\oldbm~and~\ourbm. There is a clear drop in performance for both models on~\ourbm, with both mT5 and ByT5 dropping upto 16 F1 points on the CB dataset. The mT5 model also drops by roughly 9 points in accuracy on the BoolQ dataset, and similarly 9 F1 points on the ReCoRD dataset. While the ByT5 model is in general more robust than the mT5 model, its performance also drops by 10 points in accuracy on RTE.

The spell correction baselines (Rows 3a, 3b, 4a, 4b) also fail to recover performance. With NeuSpell, mT5 sees a drop in BoolQ and RTE, slight improvement on CB, MultiRC, Record, WIC (<2 points Acc/F1). On COPA, we observe a substantial recovery (55 -> 64). For ByT5 however, there is a drop in performance across the board. NeuSpell is not well equipped to handle phonetic misspellings. Therefore the spell corrected word is often farther from the original word than the misspelling. These bad corrections hurt ByT5, which is slightly more robust to misspellings than mT5.
With Bert-Large mask prediction, for mT5 there is a slight improvement on COPA and improvement on CB(74.21 ->79.6), but worse performance on all other tasks. Again for ByT5, we see degradation in performance across the board. Since 30\% of the tokens are phonetically misspelt, the contextual mask prediction task is also not accurate. Another failure mode we observed was that the prediction is often the correct type (adjective for adjective) but not the original token.

This clearly demonstrates the challenge posed by phoneme-shift based noisy misspellings introduced in~\ourbm~. Current models and training schemes are ill-equipped to function on such data.


%% file: robust_models.tex
\section{Phoneme Prediction as a Pre-training Task}\label{sec:robust_models}
\begin{figure}[t]
    \centering
    \includegraphics[width=\linewidth]{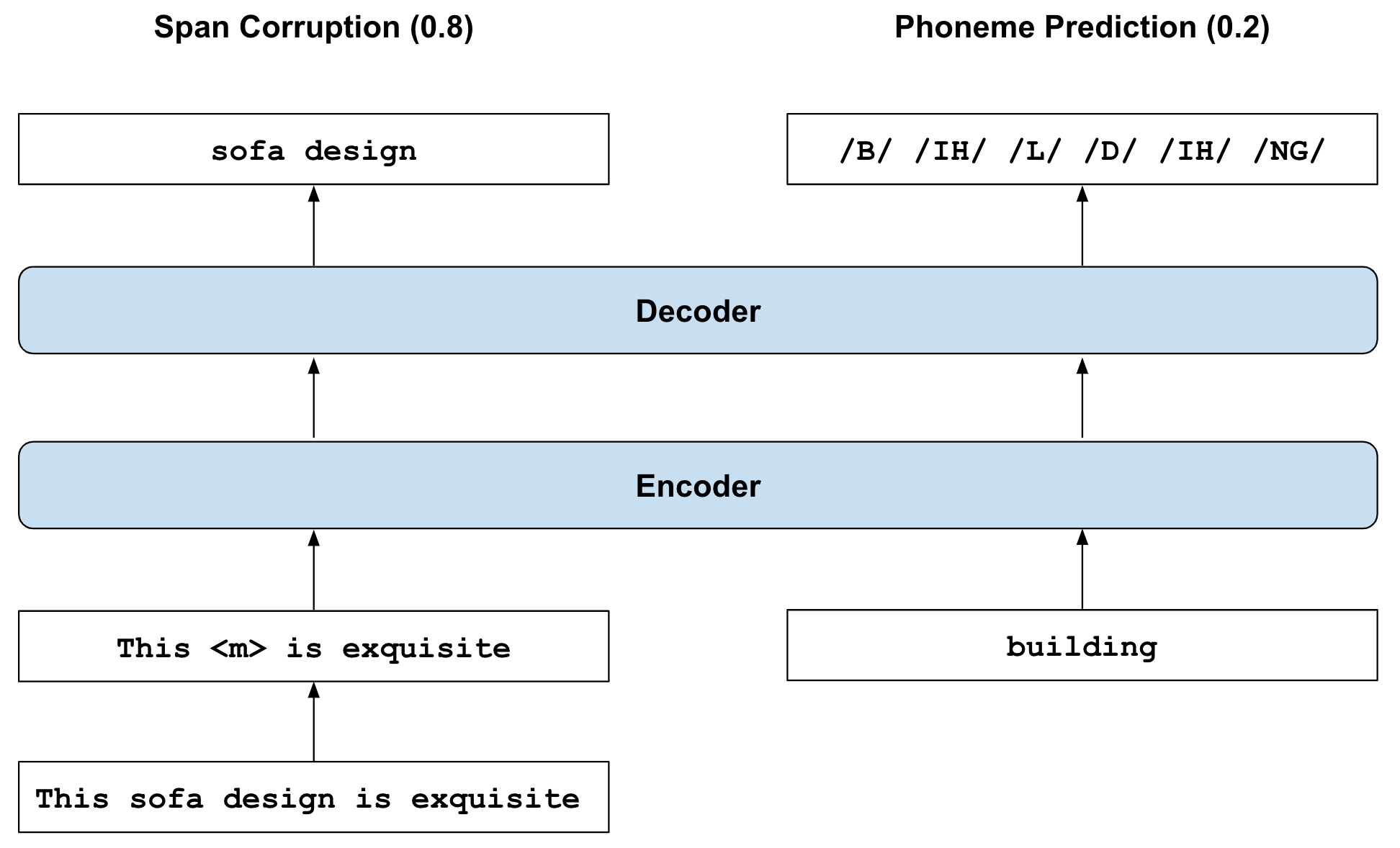}
    \caption{Demonstration of our mixture pre-training task that combines standard span-corruption with the novel phoneme prediction task in an 80:20 ratio. All weights and embeddings in the model are shared.}
    \label{fig:phoneme_pred}
\end{figure}
Given the inadequacy of existing State-of-the-Art models in handling phonetic noise in inputs, we propose a novel pre-training task of phoneme prediction. We posit that the task of predicting phoneme sequences will have the effect of teaching the model ``phonetic information". Since different lexicalizations of the same sound will have the same phoneme sequence, the model will learn to embed these close. Additionally since close sounds often appear in similar intra-word contexts, their graphemic representations will also be pushed closed together.

However, to perform NLP tasks, semantic similarity is still crucial. In current models this is often achieved through some variation of the span corruption task (corrupting a span in the input and predicting it on the output). We propose a mixture of these two tasks where a small amount of the phoneme prediction task (20\%) is mixed into the standard span corruption task. Figure~\ref{fig:phoneme_pred} demonstrates our proposal through two example instances.  In the first instance the span ``sofa design" is masked in the input (replaced with a sentinel) and is expected to be produced on the output. This teaches the model that adjectives like ``exquisite" are semantically close. The second instance has the word ``building" in the input and the phoneme sequence corresponding to this word (B, IH, L, D, IH, NG) on the output. This task teaches the model that all tokens that produce the same sound (like ``ui" or ``e" for IH) should be embedded close.

We train both mT5 and ByT5 checkpoints for an additional 100,000 steps (10\% additional steps) on this mixture task. We call this step of additional pre-training, ``Phonetic pre-training". Finally, we fine-tune these models on the standard clean~\oldbm~training set. The phoneme prediction data is created by taking roughly 2,000,000 highest frequency words from the Common Crawl English data and getting their pronunciations from an off-the-shelf Grapheme to Phoneme model. As we will see later, this kind of noisy supervision (not human labelled) is still useful in making models phonetically robust.

The last two rows in Table~\ref{tab:super_gloo_results} show the performance of these models on~\ourbm. We find that the simple additional pre-training step of phoneme-prediction substantially improves performance of the ByT5 model on the noised benchmark (row 6 against row 4). Performance on CB increases by 11 F1 points, on COPA there is a 8 point accuracy gain, and a 5 point accuracy gain on RTE. While performance still lags compared to the clean benchmark~\oldbm~(row 6 against row 2) on most tasks, for MultiRC and COPA, we find that the phonetically pre-trained ByT5 model even out-performs the vanilla pre-trained model (row 2) numbers on the clean task. This is particularly impressive because the Phonetic ByT5 model (row 6) has never seen any noisy data during its training. The mT5 model does not however see the same impressive gains through this pre-training task. We hypothesize this is because of the harder sub-word tokenization in mT5. Many tokens that this model needs on the noised task are never seen when it's trained on clean data and therefore have poor representations.

The ByT5 model does however have certain drawbacks.  Since input sequences are much longer with byte level representations, both training and inference times are much slower than a sub-word tokenized alternative (like mT5). Additionally, the byte-level representation also restricts input sequence lengths.  Using these phonetically robust byte-level models as teachers for sub-word tokenized student models remains an interesting direction for future work.